\documentclass[conference]{IEEEtran}
\IEEEoverridecommandlockouts
\usepackage{cite}
\usepackage{amsmath,amssymb,amsfonts}
\usepackage{algorithmic}
\usepackage{graphicx}
\usepackage{textcomp}
\usepackage{xcolor}
\usepackage{amsmath}
\usepackage{xurl}
\usepackage{multirow}
\usepackage{graphicx}
\usepackage{booktabs}

\def\BibTeX{{\rm B\kern-.05em{\sc i\kern-.025em b}\kern-.08em
    T\kern-.1667em\lower.7ex\hbox{E}\kern-.125emX}}
\begin{document}

\title{GluMarker: A Novel Predictive Modeling of Glycemic Control Through Digital Biomarkers
\thanks{$^*$Equal contribution. $^\dag$Corresponding author.}
}

\author{\IEEEauthorblockN{Ziyi Zhou$^{*\dag1}$, Ming Cheng$^{*1}$, Xingjian Diao$^{*1}$, Yanjun Cui$^1$, Xiangling Li$^2$}
\IEEEauthorblockA{$^1$Department of Computer Science, Dartmouth College, Hanover, NH 03755, USA \\
$^2$Department of Biomedical Engineering, Dartmouth College, Hanover, NH 03755, USA \\
Email: ziyi.zhou.gr@dartmouth.edu}
}

\maketitle

\begin{abstract}

The escalating prevalence of diabetes globally underscores the need for diabetes management. Recent research highlights the growing focus on digital biomarkers in diabetes management, with innovations in computational frameworks and noninvasive monitoring techniques using personalized glucose metrics. However, they predominantly focus on insulin dosing and specific glucose values, or with limited attention given to overall glycemic control. This leaves a gap in expanding the scope of digital biomarkers for overall glycemic control in diabetes management. To address such a research gap, we propose GluMarker -- an end-to-end framework for modeling digital biomarkers using broader factors sources to predict glycemic control. Through the assessment and refinement of various machine learning baselines, GluMarker achieves state-of-the-art on Anderson's dataset in predicting next-day glycemic control. Moreover, our research identifies key digital biomarkers for the next day's glycemic control prediction. These identified biomarkers are instrumental in illuminating the daily factors that influence glycemic management, offering vital insights for diabetes care.

\end{abstract}
\newcommand\blfootnote[1]{%
  \begingroup
  \renewcommand\thefootnote{}\footnote{#1}%
  \addtocounter{footnote}{-1}%
  \endgroup
}

\blfootnote{46th Annual International Conference of the IEEE Engineering in Medicine \& Biology Society, 2024}

\section{Introduction}
Digital biomarkers \cite{dagum2018digital,mahadevan2020development} are an emergent class of medical indicators derived from data collected via digital devices, such as wearables, smartphones, and other sensor-equipped technologies. Unlike traditional biomarkers that are typically obtained from biological samples in a clinical setting\cite{babrak2019traditional, hall2011risk}, digital biomarkers enable the quantification of physiological, behavioral, and environmental factors, offering a granular view of an individual's health status \cite{kourtis2019digital,babrak2019traditional,bartolome2022computational}. The integration of digital biomarkers in medical research and practice is opening avenues for significant progress in areas like chronic disease management \cite{kourtis2019digital, wang2024differential}, smart healthcare \cite{wang2020guardhealth}, diabetes treatment systems \cite{li2021fully, li2022smart}, and the development of wearable technologies~\cite{jacobson2020digital, AlikhaniKoshkak2024SEAL}.

Diabetes, a complex metabolic disorder characterized by persistent hyperglycemia \cite{egan2019diabetes}, has become a major global health concern. As diabetes affects approximately 537 million people worldwide~\cite{9871646}, the importance of effective glycemic control ~\cite{bartolome2022computational} is increasingly crucial in managing this growing health challenge~\cite{saeedi2019global}.

Digital biomarkers in diabetes are increasingly gaining attention in recent medical research and practice \cite{bent2021engineering,bartolome2022computational}. Bartolome et al.~\cite{bartolome2022computational} present a computational framework for identifying digital biomarkers to manage glycemic control in diabetes, using data from continuous glucose monitors (CGMs) and insulin pumps. Bent et al.~\cite{bent2021engineering} introduce an innovative approach to noninvasively monitor and predict interstitial glucose levels using data from smartwatches and food logs. A significant contribution of this study \cite{bent2021engineering} is the establishment of personalized definitions of glucose excursions (PersHigh, PersLow, PersNorm) based on individual blood glucose measurement variations.

Despite the increasing interest, the field of digital biomarkers in diabetes is still in its early stages, possessing substantial potential that remains largely unexplored.
Part of previous research has primarily focused on insulin dosing data (from insulin pumps) and blood glucose data (from CGMs) \cite{zhou2024crossgp}, while others generally concentrate on predicting short-term blood glucose fluctuations \cite{cheng2024shortterm} instead of addressing the broader aspects of overall glycemic control.

This results in a \textit{\textbf{notable gap}} in understanding and identification of a broader range of biomarkers for overall glycemic control, underscoring the need for a more comprehensive exploration of potential digital biomarkers in diabetes.
Addressing these challenges is essential for both diabetes management and broader medical research~\cite{zhou2021doseguide,10124956,cheng2023saic,zhang2023doseformer, shi2023prompt, pedram2023experience}, enhancing the scope and effectiveness of digital biomarkers in clinical applications. 
Therefore, one crucial question has arisen: \textbf{\textit{How can we broaden the investigation of digital biomarkers for overall glycemic control in diabetes management?}}

To overcome the challenge above, we propose \textbf{GluMarker} -- an end-to-end pipeline for modeling digital biomarkers utilizing broader factors sources to predict glycemic status (i.e., good, moderate, or poor) of the next day.

In summary, our contribution is threefold:
\begin{enumerate}
    \item {
    \textbf{Exploration on broader factors sources.} We incorporate a broader range of data sources such as meal size in addition to CGMs and insulin pump data.
    }
    \item {
    \textbf{State-of-the-art performance.} 
     GluMarker achieves state-of-the-art on Anderson's dataset \cite{anderson2016multinational} in predicting glycemic control of the next day.
    }
    \item {
    \textbf{Effective digital biomarkers identification.} 
    Our study extensively investigates digital biomarkers, identifying those with the ability to predict the next day's glycemic control accurately. These identified biomarkers are crucial for understanding daily variables that affect glycemic management, thus providing indispensable insights for effective diabetes care.
    }
\end{enumerate}

\section{Methods}

\subsection{Digital Biomarker Modeling}
\label{sec:digital}
Assume the set $\mathcal{X} = \{X_1, X_2, ..., X_n\}$ indicates the group of $n$ features where $X_i$ represents each feature vector. 
Based on the data distribution, we divide the feature vector $X_i$ into several intervals, following \cite{bartolome2022computational}.
Formally, suppose $X_i$ is divided into $m$ intervals: $X_i = \{I_1, I_2, ..., I_m\}$, and each interval $I_k = [a_k, b_k)$ has the lower bound and upper bound of $a_k$ and $b_k$, respectively. Therefore, the relationship between each interval can be expressed as:
\begin{equation}
    a_k < b_k < a_{k+1} < b_{k+1}, \quad k \in [1, m-1]
\end{equation}

Let  $\mathcal{B} = \{B_1, B_2, ..., B_m\}$ denote the set of digital biomarkers derived from the feature vector $X_i$. Each digital biomarker $B_j$ is associated with the corresponding interval $I_j$ and is computed based on the aggregation of the values falling within that interval.
Therefore, $\mathcal{B}$ can be utilized as digital biomarkers representing feature $X_i$. 

As discussed in \cite{bartolome2022computational}, we apply the features of two days (present and past) to enhance glycemic control.
The demonstration of digital biomarker modeling on Anderson's dataset \cite{anderson2016multinational} is shown in Section \ref{exp:digital}.

\subsection{GluMarker -- Ours}

\begin{figure}[tb]
  \centering
  \includegraphics[width=0.7\linewidth]{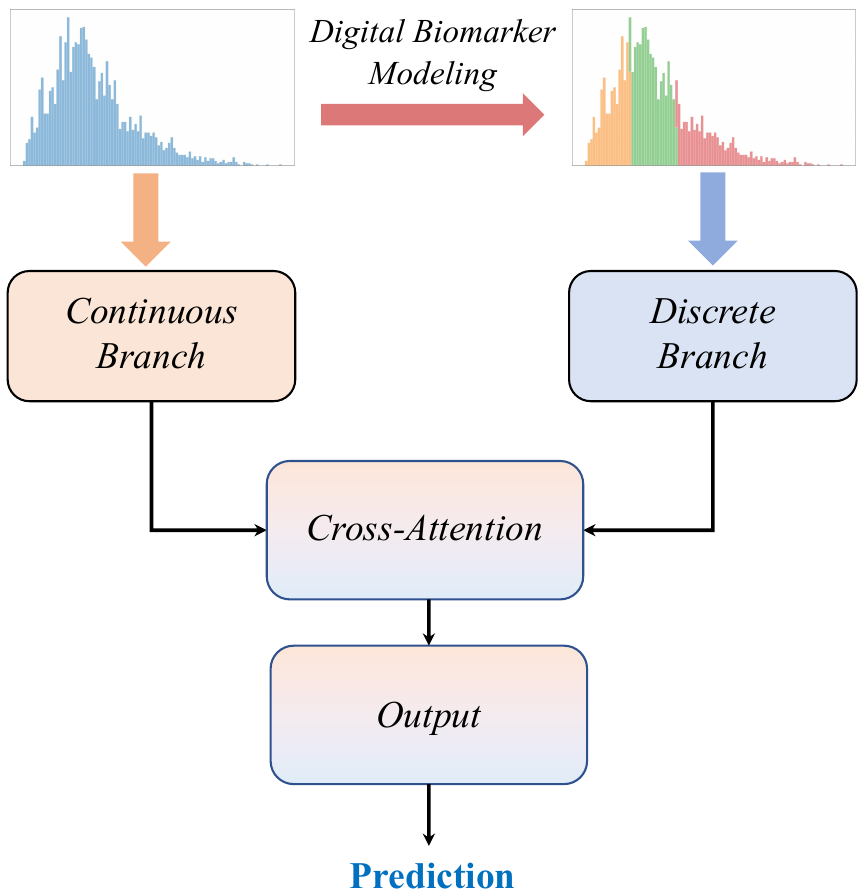}
  \caption{\textbf{
  Architecture of GluMarker.
  } 
  The original data is divided into intervals to create digital biomarkers, which are then input into the model for glycemic control prediction. 
  }
  \label{fig:model}
\end{figure}

To achieve accurate glycemic prediction based on the input features, a novel machine-learning model is designed, as expressed in Figure \ref{fig:model}.
Since the procedure in Section \ref{sec:digital} transforms the continuous feature vectors into discrete ones, we design a parallel-branch architecture to extract continuous/discrete features, respectively. 
Specifically, the continuous branch $B_c$ accepts the original data as feature input, and outputs the feature representations through sequential convolutional blocks (Multi-layer Dense blocks). Similarly, the discrete branch $B_d$ takes discrete digital biomarker features as input, and learns corresponding representations through convolution operations (Multi-layer Dense blocks with a different depth). The two operations can be mathematically expressed as:  
\begin{equation}
    R_c = B_c(F_c), \quad R_d = B_d(F_d)
\end{equation}
where $F_c$ and $F_d$ indicate the continuous and discrete features, 
$R_c$ and $R_d$ are the output representations by these two modules, 
$B_c$ and $B_d$ represent the continuous and discrete branches, respectively. 

Inspired by the relevant work discussing the cross-attention mechanism \cite{diao2023av, diao2023ft2tf, yang2022gammae, jian2024bootstrapping} for various domain-adaptation \cite{de2021adversarial, hu2020dasgil, kasahara2018unsupervised} representation learning tasks, representations from two domains ($R_c, R_d$, continuous/discrete domain) are then fused through an attention layer with sigmoid activation function \cite{han1995influence} to instruct the model to selectively assign adaptive weights and capture the main components from the features. The final predictions $pred$ are then generated through a fusion output layer $O_f$ (single-layer Dense block), as expressed below:
\begin{equation}
    pred = O_f(Attn(R_c, R_d))
\end{equation}

Following the standard criterion \cite{zhang2018generalized, mao2023cross}, we employ cross-entropy loss to train the model. 

Benefiting from the parallel-branch design, our model implements feature fusion from two domains to enhance feature aggregation, making accurate glycemic control predictions of diabetes patients based on the digital biomarker strategy. 

\begin{figure*}[tb]
  \centering
  \includegraphics[width=\linewidth]{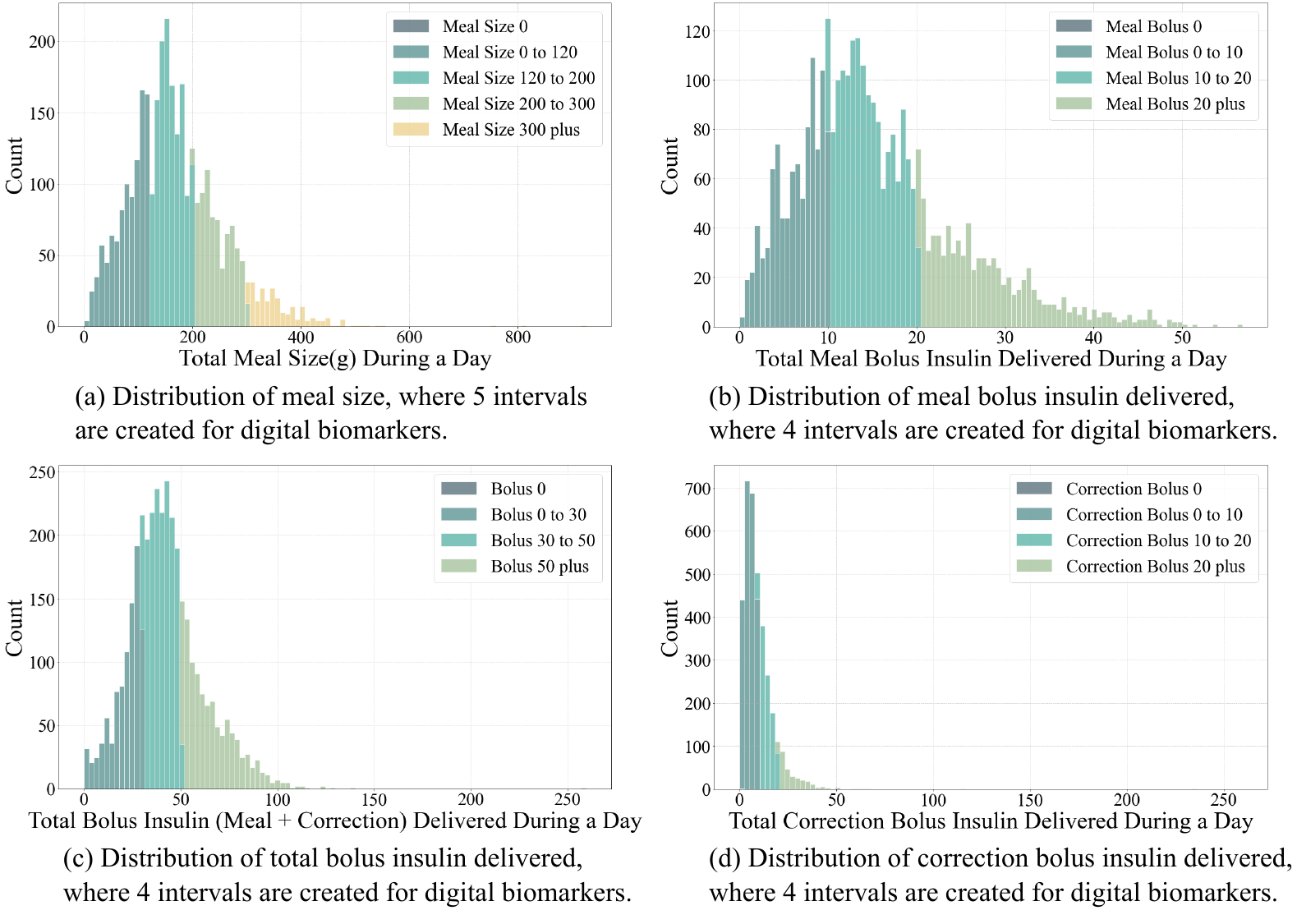}
  \caption{\textbf{
  Visualization of data distribution. 
  } 
  Based on the data distribution, multiple margins are generated to divide the data into several intervals, with each interval representing a digital biomarker.
  }
  \label{fig:digitalbio}
\end{figure*}

\begin{figure}[tb]
  \centering
  \includegraphics[width=\linewidth]{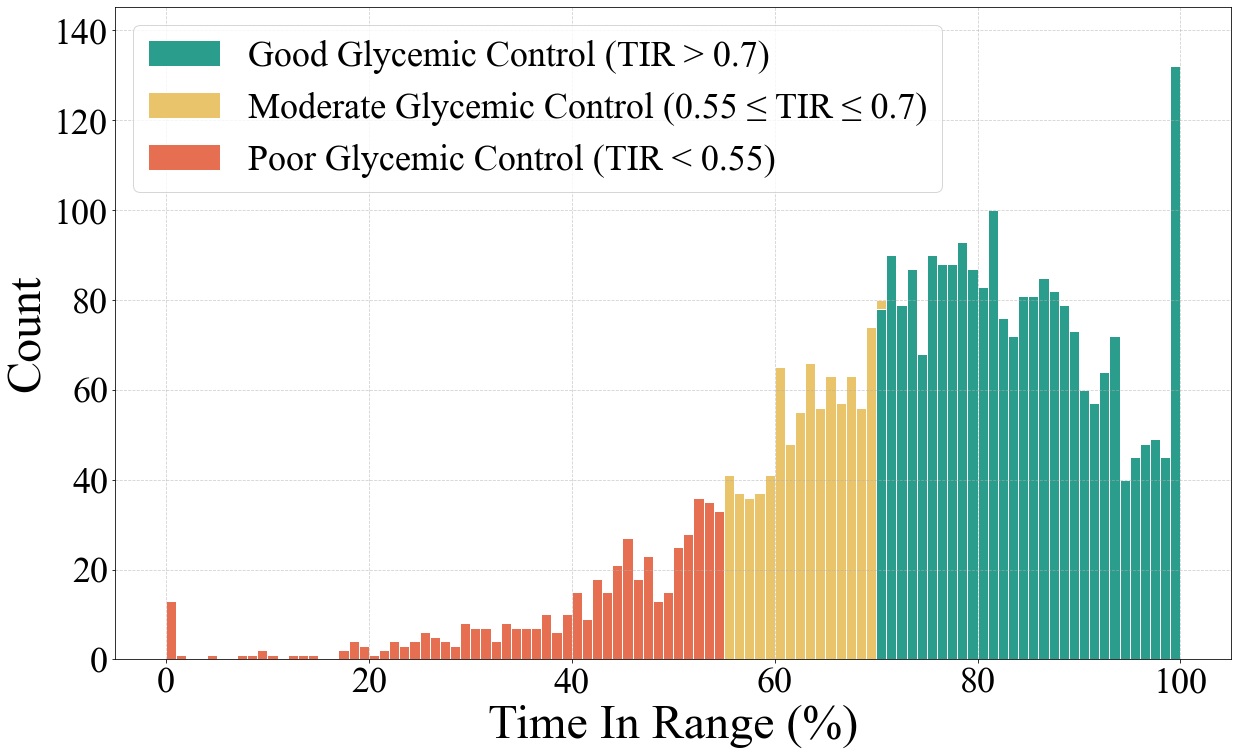}
  \caption{\textbf{
  Visualization of three glycemic control situations.
  } 
  Following Equation \ref{controlrange}, we generate three glycemic control ranges for the prediction task. 
  }
  \label{fig:cgm}
\end{figure}

\subsection{Comparison Methods}
We employed the Receiver Operating Characteristic (ROC) curve as the primary assessment metric. Furthermore, we established a comparative analysis using Linear Support Vector Classification (Linear SVC) \cite{cortes1995support}, Naive Bayes \cite{murphy2006naive}, and Multilayer Perceptron (MLP) \cite{popescu2009multilayer} as baseline models.

\subsubsection{ROC Curve}

The Area Under the ROC Curve (AUC) allows us to assess the model's ability to discriminate glycemic status accurately in an intuitive way. The AUC ranges from 0 to 1, where:
\begin{itemize}
    \item An AUC of 1 indicates a perfect model that perfectly discriminates between the classes.

    \item An AUC of 0.5 suggests a model with no discriminative ability, equivalent to random guessing.

    \item An AUC less than 0.5 suggests worse-than-random predictions, indicating serious issues in the model or data.
\end{itemize}

\begin{figure*}[tb]
  \centering
  \includegraphics[width=\linewidth]{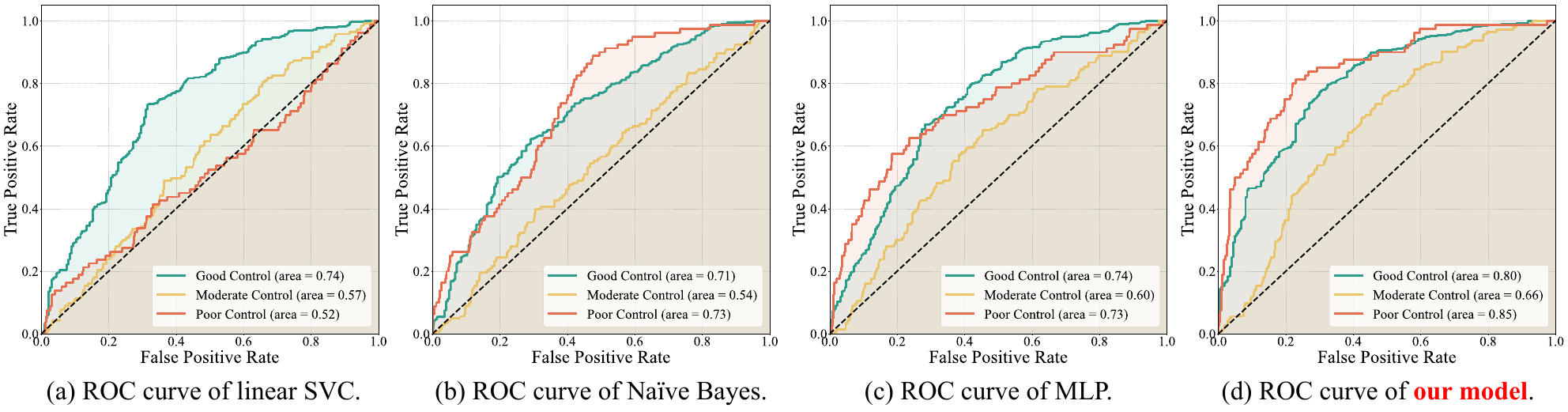}
  \caption{\textbf{
  Visualization of ROC curves generated by different models. 
  } 
  Our model produces the largest area under the ROC curve (AUC score), indicating the best classification performance and highest sensitivity, and the lowest false positive rate.
  }
  \label{fig:roc}
\end{figure*}

\subsubsection{Baseline models} 
To establish a robust evaluation, we selected three baseline models with distinct characteristics, providing a broad spectrum of comparison for our model.

\begin{itemize}
    \item Naive Bayes: Naive Bayes classifiers are a family of simple probabilistic classifiers based on applying Bayes' theorem with strong (naive) independence assumptions between the features. The choice of Naive Bayes is motivated by its efficiency, simplicity, and performance in tasks with conditional independence between features. It provides a probabilistic perspective, beneficial for understanding feature contributions in glycemic control prediction.

    \item Linear Support Vector Classification (Linear SVC): Linear SVC algorithm constructs a hyperplane in the feature space that maximally separates the classes with a margin. Given its effectiveness in high-dimensional spaces, Linear SVC serves as a strong baseline to evaluate the linear discriminative power of the features in distinguishing between glycemic control categories.

    \item Multilayer Perceptron (MLP): MLP is a class of feedforward artificial neural networks that consists of at least three layers of nodes: an input layer, a hidden layer, and an output layer. MLP was chosen for its capability to model non-linear relationships and interactions between features. Its inclusion as a baseline allows us to assess the advantage of deep learning techniques in capturing complex patterns in the data related to glycemic control.
    
\end{itemize}

\begin{figure*}[tb]
  \centering
  \includegraphics[width=\linewidth]{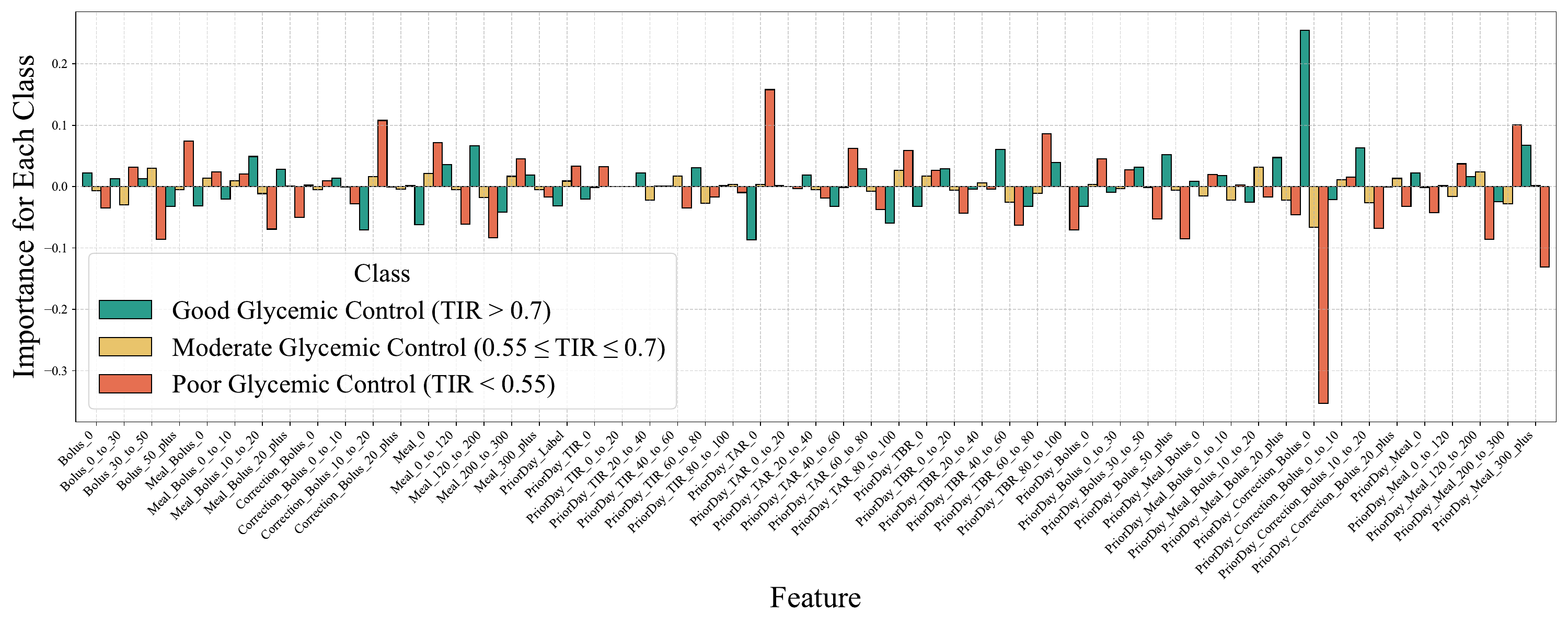}
  \caption{\textbf{
  Visualization of feature importance for glycemic control prediction. 
  } 
  The impact of all features for each category (``Good", ``Moderate", and ``Poor") is demonstrated. For instance, the prior-day correction bolus and TAR play the most significant roles in good and poor glycemic control, respectively. 
  }
  \label{fig:featureimp}
\end{figure*}

\begin{figure*}[tb]
  \centering
  \includegraphics[width=0.8\linewidth]{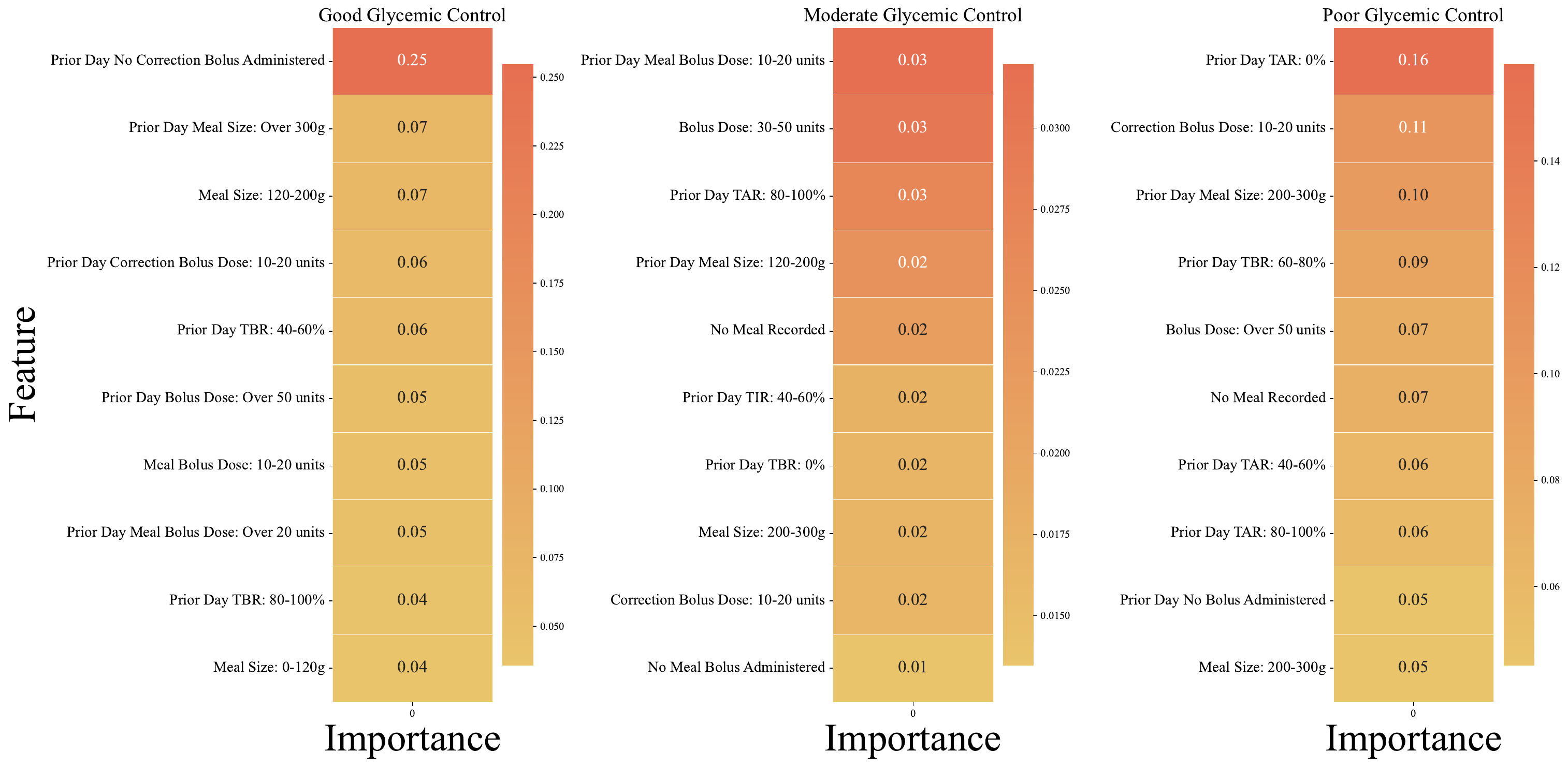}
  \caption{\textbf{
  Feature impact heatmap demonstration.
  } 
  \textit{Top-10} important features for three categories (good, moderate, and poor) prediction are shown.
  }
  \label{fig:featureImpHeat}
\end{figure*}

\section{Experiments}
\subsection{Dataset Description and Pre-processing}
We conduct extensive experiments on Anderson's dataset \cite{anderson2016multinational} containing the real records of 30 diabetics to thoroughly and objectively analyze the model performance. 
To transfer original features into digital biomarker forms discussed in Section \ref{exp:digital}, inappropriate and unrelated features (e.g. gender) are ignored. Therefore, 
we only involve a subset of features (related to meal intake and insulin dose) in the experiment: ``CGM", ``TotalBolus", ``TotalMealBolus", ``TotalCorrectionBolus", ``TotalMealSize". Inspired by \cite{bartolome2022computational}, features recorded on both the previous day and the present day are included for model training.

Since the original blood glucose records change frequently, we divide the blood glucose values (``CGM" feature) into three intervals to get relatively stable features: TIR (time-in-range), TAR (time-above-rang), and TBR (time-below-range), following \cite{bergenstal2013recommendations}.

To clearly and concisely present the level of patients' glycemic control, we categorize results into 3 ranges: ``Good", ``Moderate", and ``Poor", reflecting different control situations, respectively. Specifically, the three ranges are constructed following the equation below: 
\begin{equation}
\label{controlrange}
\begin{cases}
  \text{Good Control} & \text{TIR} > m_{u}\\
  \text{Moderate Control} & m_{l} \leq \text{TIR} < m_{u} \\
  \text{Poor Control} & \text{TIR} < m_l
\end{cases}      
\end{equation}
where $m_l = 0.55$ and $m_u = 0.7$ denote the lower/upper bound (percentage of time) \cite{bartolome2022computational}, respectively. The visualization of three glycemic control situations is shown in Figure \ref{fig:cgm}. 

Therefore, the glycemic control prediction task is transferred into a 3-classification task, in which the model will predict the expected control range.

\subsection{Digital Biomarkers Generation}
\label{exp:digital}

The illustration of digital biomarkers is shown in Figure \ref{fig:digitalbio}, where we divide original features into several intervals based on their distribution. For example, we divide ``meal size" into 5 intervals: no-entry, 0-120 units, 120-200 units, 200-300 units, and greater than 300 units, with each interval representing a different and unique meal intake condition. 
Similarly, the ``total bolus insulin" is divided into 4 intervals: no-entry, 0-30 units, 30-50 units, and greater than 50 units. These intervals can respond to different levels of medical interventions and thus serve as digital biomarkers. 

Afterward, the digital biomarkers are then combined with original records as the model input for glycemic control prediction.

\subsection{Glycemic Control Prediction}
To evaluate our model's performance comprehensively, we implement three baselines (Linear SVC, MLP, Naïve Bayes) for comparison, with the quantitative results shown in Figure \ref{fig:roc}.
It is obvious that our model shows the largest area under the ROC curve (AUC score), indicating the best classification performance. Specifically, our model represents an AUC of $0.80$ (``Good Glycemic Control"), $0.66$ (``Moderate Glycemic Control"), and $0.85$ (``Poor Glycemic Control"). Considering that diabetic patients generally lack good glycemic control, the high prediction performance ($0.85$ of AUC score, red line in the figure) of our model for the ``Poor" category specifically demonstrates the potential to assist in the clinical treatment of diabetics in real-life situations.

\subsection{Feature Importance For Glycemic Control}
We develop a comprehensive methodology to evaluate the impact of individual features on the prediction accuracy across different glycemic control categories. Our approach hinges on the principle of isolating the effect of each binary feature by artificially setting its value to $1$ while keeping all other features constant. This experimental manipulation allows us to directly observe the change in predictive probabilities for each category, offering insights into how each feature influences our model's decision-making process.

In the evaluation process, we create two versions of the dataset for each binary feature: one with the feature of interest set to $1$ for all instances, and another as the original dataset. We then feed both versions into our pre-trained model and calculated the difference in average predictive probabilities across the three glycemic control categories respectively. The calculated differences quantify the relative importance of each feature in predicting the outcome categories.

The importance of all features for diabetics' glycemic control is shown in Figure \ref{fig:featureimp}, where we visualize the effect of each feature on the model's predictions. For example, the correction bolus of the prior day (the highest green bar) plays the most significant role in good glycemic control, while the TAR of the prior day (the highest red bar) has the greatest impact on poor glycemic control predictions. 
The above analysis results can assist researchers in selecting certain features for more accurate prediction of glycemic control in subsequent studies.

\subsection{\textit{Top-10} Digital Biomarkers Visualization}
The features selected by our model in this analysis are potential biomarkers for glycemic control because they encapsulate critical aspects of diabetes management, including insulin dosage (bolus and correction), meal intake, and prior-day glycemic metrics.

The \textit{top-10} digital biomarkers generated by our model are demonstrated in Figure \ref{fig:featureImpHeat}, 
representing the 10 digital biomarkers that best characterize diabetics' glycemic control.  

We noted that no correction bolus on the previous day was the most important digital biomarker of good glycemic control. This is consistent with common sense since no correction bolus on the prior day generally means that glucose remains good and no additional doses are needed; in this case, there is a greater probability of maintaining good glycemic control the next day. Surprisingly, having a meal larger than 300g on the prior day is also considered by the model to be an important indicator of good glycemic control on the next day. We believe that this is because this indicator has a strong correlation with the indicator that the Meal Bolus was greater than 20 units the previous day, and the two often appear at the same time. This result shows that if the patient manages glucose well during meals the day before, then there is a greater chance of maintaining good glycemic control the next day.

In addition, the 
prior-day meal bolus dose ($10-20$ units), insulin bolus ($30-50$ units) and the prior-day TAR ($80-100\%$) rank among the prominent digital biomarkers exhibiting the highest likelihoods linked to moderate glycemic control. 
Similarly, a combination of prior-day TAR ($0\%$) and present-day correction bolus dose ($10-20$ units) stand out as the key digital biomarker associated with elevated probabilities of poor glycemic control.

An interesting observation from our model's analysis is the significant predictive relationship between a Total Above Range (TAR) of 0\% on the preceding day and an increased likelihood of deteriorated glycemic control on the subsequent day. This outcome suggests that within our dataset, a continuity of optimal glycemic control across consecutive days is relatively uncommon. This finding underscores a critical message for patients with diabetes: the achievement of satisfactory glycemic metrics on one day should not lead to complacency regarding management strategies for the following day. It highlights the necessity for consistent vigilance and adherence to management protocols to maintain glycemic control over time.

The above results provide effective and critical guidance for the clinical treatment of diabetics, which can be further generalized to be applied in the real world.

\section{Conclusion}

The paper introduces GluMarker, a novel framework designed to predict next-day glycemic control through digital biomarkers from a wide range of data sources, achieving state-of-the-art prediction accuracy on Anderson’s dataset. Additionally, the study identifies effective digital biomarkers, deepening insights into daily factors that influence glycemic management. Future work will aim to enhance GluMarker's predictive capabilities by integrating additional data (e.g. stress, exercise, and behavioral habits) and examining its effectiveness across various patient populations to expand its application in personalized diabetes care.

\vspace{12pt}

\bibliographystyle{IEEEtran}

\bibliography{bibtex/IEEEexample}

\end{document}